# AI enhanced finite element multiscale modelling and structural uncertainty analysis of a functionally graded porous beam

Da Chen[1234]*, Nima Emami[2], Shahed Rezaei[2]*, Philipp L. Rosendahl[3], Bai-Xiang Xu[2], Jens Schneider[3], Kang Gao[5], Jie Yang[6]

[1] *Department of Infrastructure Engineering, The University of Melbourne, VIC 3010, Australia*
[2] *Mechanics of Functional Materials Division, Department of Materials Science, Technical University of Darmstadt, Germany*
[3] *Institute of Structural Mechanics and Design, Department of Civil and Environmental Engineering, Technical University of Darmstadt, Germany*
[4] *School of Civil Engineering, The University of Queensland, St. Lucia, QLD 4072, Australia*
[5] *School of Civil Engineering, Southeast University, No.2 Dongnandaxue Road, Jiangning Disctrict. 211189, China*
[6] *School of Engineering, RMIT University, PO Box 71, Bundoora, VIC 3083, Australia*
* Corresponding author: da.chen1@unimelb.edu.au (Chen) s.rezaei@mfm.tu-darmstadt.de (Rezaei)

**Abstract**

The local geometrical randomness of metal foams brings complexities to the performance prediction of porous structures. Although the relative density is commonly deemed as the key factor, the stochasticity of internal cell sizes and shapes has an apparent effect on the porous structural behaviour but the corresponding measurement is challenging. To address this issue, we are aimed to develop an assessment strategy for efficiently examining the foam properties by combining multiscale modelling and deep learning. The multiscale modelling is based on the finite element (FE) simulation employing representative volume elements (RVEs) with random cellular morphologies, mimicking the typical features of closed-cell Aluminium foams. A deep learning database is constructed for training the designed convolutional neural networks (CNNs) to establish a direct link between the mesoscopic porosity characteristics and the effective Young's modulus of foams. The error range of CNN models leads to an uncertain mechanical performance, which is further evaluated in a structural uncertainty analysis on the FG porous three-layer beam consisting of two thin high-density layers and a thick low-density one, where the imprecise CNN predicted moduli are represented as triangular fuzzy numbers in double parametric form. The uncertain beam bending deflections under a mid-span point load are calculated with the aid of Timoshenko beam theory and the Ritz method. Our findings suggest the success in training CNN models to estimate RVE modulus using images with an average error of 5.92%. The evaluation of FG porous structures can be significantly simplified with the proposed method and connects to the mesoscopic cellular morphologies without establishing the mechanics model for local foams.

# 1. Introduction

The development of lightweight structures is an active research field for providing solutions to crack the long-term dilemma between weight and stiffness. The recent innovations, such as carbon fibre reinforcing systems [1] and new lightweight hybrid materials [2], are increasingly appreciated in the industrial practice, where the light yet stiff components are critically applied in the production of vehicle frames [3], aircraft skeletons [4, 5], and spatial building designs [6, 7], etc.

The porous metal-matrix structures are among the most adopted lightweight forms in structural and mechanical engineering, and involve honeycombs, lattices, and foams in line with the geometrical characteristics. Compared to honeycombs [8] and lattices [9] containing two dimensional (2D) or three dimensional (3D) regularised cellular morphologies, metal foams are featured by random distributions of internal pore sizes, densities, shapes, and consequently relatively low manufacturing costs. In addition to the significantly reduced weight (occasionally over 90% lighter), they are also underlined with notable multifunctional properties, including definable density, good energy absorption, and useful thermal and acoustic traits [10]. Integrating varying metal foams into one single structural component gives rise to the FG porous structures (metal foam based), which have been demonstrated to possess even more outstanding mechanical attributes [11-16].

Despite the speedy advance in the research of FG porous structures, predictions of their properties have been continuously intricate, induced from the random nature of metal foams. Although the relative density is widely regarded as a dominant factor, the contribution from cell wall arrangements can be significant and changing stochastic porosity distributions may result in evident variances when determining the overall foam mechanical features. For example, Young's moduli of closed-cell Aluminium foams, discussed by Chen et al. [17], have a distinct variation up to 14.38 % under the same relative density, matrix material, wavy edge, and RVE size. This inherent difference is difficult to be directly identified from the foam morphologies.

On the other hand, the rapid development of AI enhanced engineering provides us a powerful tool to resolve problems challenging for conventional approaches [18-20], and is promising to open up new horizons for cracking the aforementioned issue in the evaluation of porous structures made of random foams. Among various AI approaches, deep learning [21-23] has gained much attention of different research societies in the last decade and has been widely used in numerous applications. One of the factors that helped this outreach is the massive amount of data available in each field. With the high computation power and the advances in simulation and modelling of the physical phenomenon, researchers have considered generating synthetic data to train models, which was not possible earlier due to a lack of data or severe cost and resources required for acquiring those data. Deep learning models could be considered as a universal solution for data-driven problems meaning these models can handle all types of data. Deep learning models such as recurrent neural networks (RNNs) [24], CNNs [25], and other models can be incorporated into various applications based on the nature of data and desired outputs.

This work adopts deep learning to measure how the stochastic porosity distributions impact the metal foam behaviour with a focus on the influence of random cell wall arrangements, exploring a novel way to calculate the foam properties directly from cellular morphologies. We are aimed to offer solutions for effectively examining porous structures simply according to their images without building up the corresponding mechanics models. Specifically, CNNs are employed here to investigate Young's modulus of the porous materials, and to establish the essential structure-property relation as efficiently as possible. CNNs provide significant insights into the correlation between images and the features they represent, and have been widely adopted in various fields, i.e., image segmentation, investigating the microstructure-property relations, detection and prediction of fault and failure [26, 27]. Following the deep learning, the structural uncertainty analysis is performed on the static bending response of an FG porous three-layer beam, in order to identify its deflection ranges considering the indeterministic CNN estimated properties of different layers.

## 2. AI enhanced finite element multiscale modelling

In this section, the multiscale modelling using 2D FE homogenisation is introduced (2.1) with a focus on the details of RVE calculations, followed by the establishment of database (2.2) which is used in the training of CNNs (2.3). The results predicted by deep learning models are subsequently discussed (2.4).

### 2.1. RVEs for FE homogenisation

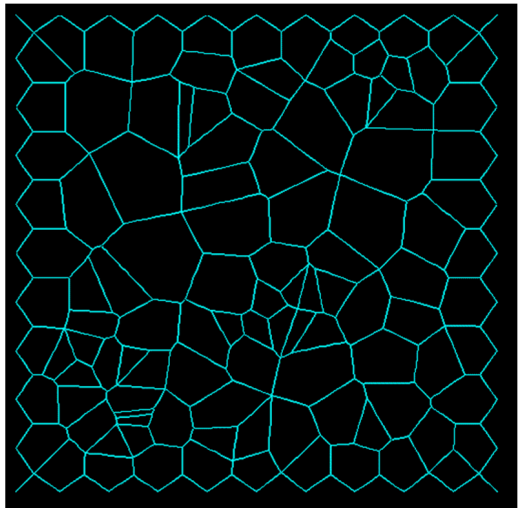

**Fig. 1.** An example of 2D square FE RVEs (image used for training CNNs).

Homogenisation is a commonly adopted approach for multiscale modelling, assuming that the overall mechanical constants of a small representative material region, i.e., RVEs, are equivalent to those of anywhere else [28-30]. Thus, regardless of the irregularity in microscopic/mesoscopic views, the general material form may be regarded as homogenised and share the features of RVEs. In this study, we employ multiscale modelling to investigate the Young's modulus of closed-cell metal foams that is adopted in the porous structural analysis. Using Voronoi tessellation where the seed positions are generated using the standard uniform distribution, 2D square FE RVEs can be constructed by stochastic porosities via commercial software MATLAB (version: R2021a) and ANSYS/APDL (version: 2020 R2), as illustrated in Fig. 1 for a typical example. Note that the wavy edge is designed for the implementation of periodic boundary conditions (PBCs) requiring the corresponding node pairs on the opposite edges to move equivalently in both horizontal and vertical directions. The internal cell wall morphologies and thicknesses jointly determine the RVE stiffness, hence the influence of regularity in edges may be ignored.

The parameters of RVEs are computed with metal foam characteristics, and the following approach is merely suitable for high-porosity foams. The estimated RVE internal cell number $N_{cell}$ is linked to the foam average cell size $\emptyset_{cell}$, which normally comes from the manufacturers, via

$$N_{cell} \approx (H/\emptyset_{cell})^2 \tag{1}$$

where $H$ denotes the height/width of square RVEs.

The total cell wall length $L_{cell}$ differs based on the morphology of each RVE and is essential in the determination of the cell wall thickness (uniform for all the cell walls)

$$t_{cell} = S_{RVE}\mu/L_{cell} \tag{2}$$

of which $S_{RVE} = H^2$ and $\mu$ is the relative density.

Following the construction of RVE models, FE simulations are performed in ANSYS to evaluate their static structural behaviours by meshing the geometry with Smart Size (1) (fine mesh, ANSYS automatic element sizing) and beam element (BEAM188). All of the out-of-plane displacements are restricted, since 2D FE homogenisation is carried out in this study with a focus on the in-plane deformations. It has been demonstrated that 2D modelling is feasible for porous structural analysis [31, 32] and closed-cell metal foam simulations [15, 33, 34]. PBCs are applied on the related RVE node pairs located along left/right edges which share the same vertical coordinates, while fixed supports are imposed across bottom edge nodes. A uniform in-plane vertical nodal displacement (0.01 × $H$, moving downwards) and zero horizontal movement are prescribed on the top, and then the reaction force $F$ is predicted from FE analysis to measure the compressive modulus of RVEs according to

$$E = F/(0.01Ht_{cell}) \tag{3}$$

that is assumed to be the Young's modulus of target metal foams.

A validation case is conducted to confirm the effectiveness of the proposed RVE modelling to mimic the foam materials in reality. We use the specimen information from an existing paper [35] focused on typical ALPORAS closed-cell Aluminium foams: specimen size 50 × 50 × 30 mm³; average cell size 2.88 mm; cell wall materials: Young's modulus 61.7 GPa, mass density 2700 kg/m³, Poisson's ratio 0.3; relative density 8.52% ± 0.74%. The corresponding RVE model is shown in Fig. 2 with external size 30 × 30 mm², cell number $N_{cell}$ = 109 (Eq. (1)), relative density 8.52%, cell wall thickness 0.1142 mm (Eq. (2)), and the same cell wall material properties as those from the experimental specimen. The predicted compressive modulus $E$ = 1.103 GPa (Eq. (3)) is within the suggested range 1.0 - 1.2 GPa [35], therefore verifying the validity of the proposed method. Please note that the RVE external sizes do not evidently impact the overall elastic stiffness when sufficient cells are enclosed [17].

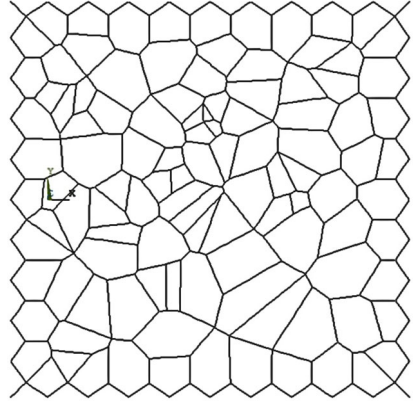

**Fig. 2.** RVE model for the validation case.

## 2.2. Database construction

Various RVEs with different internal porosity distributions and relative densities are generated and calculated to obtain their Young's moduli through FE compression simulations. The combination of RVE images (see Fig. 1 as an example) and their compressive moduli forms the database for AI training. All the RVEs adopt the same settings: size 30 × 30 mm²; BEAM188 element with the uniform thickness as 0.1142 mm; aluminium cell walls (aligned with the validation case): Young's modulus 61.7 GPa, mass density 2700 kg/m³, Poisson's ratio 0.3. Their relative densities change from 6.1 % to 20.1 %, leading to a wide value range of Young's modulus: 282.5 ~ 4587.3 MPa. During the calculations, the beam thickness is not considered as a variable when creating the database, in order to fully focus on the impact of internal morphologies. Most of the RVEs are featured with only convex cells, while some of them contain very limited numbers (normally just one) of concave cells, as illustrated in Fig. 3. In terms of the high-porosity metal foams, convex cells represent the typical internal pores, while concave cells denote those with bent or missing walls, leading to a significant drop in the compressive stiffness. Additionally, we also have included special cases highlighted by regularised cellular geometries to promote the diversity of our database. Therefore, a mix of usual (convex and random porosities, most of the cases) and unusual (concave or regular porosities) RVEs are produced, as illustrated in Figs. 3 and 4.

This dataset consists of a total of 2051 images (each linked to an individual RVE modulus) which are split into training, validation, and testing subsets with 70%, 20%, and 10%, respectively. We will show later on, that this number of images results in an acceptable range of errors in the prediction for this particular problem. Certainly, providing more image samples for the AI algorithm is indeed useful to reduce the error of predictions. The original size of the images is 512 × 512 pixels. To achieve better generalization and avoid overfitting, augmentation is introduced to the model. Since the overall material behaviour is isotropic, each image can be rotated 90°, three times, and can also be flipped both vertically and horizontally while maintaining the equivalent effective Young's modulus. Thus, when feeding an image to the model, it has a 25.0 % chance for each rotation (0°, 90°, 180°, 270°) and a 33.3 % chance of being flipped either horizontally or vertically, which helps the training model for larger epochs and more generalised applicability.

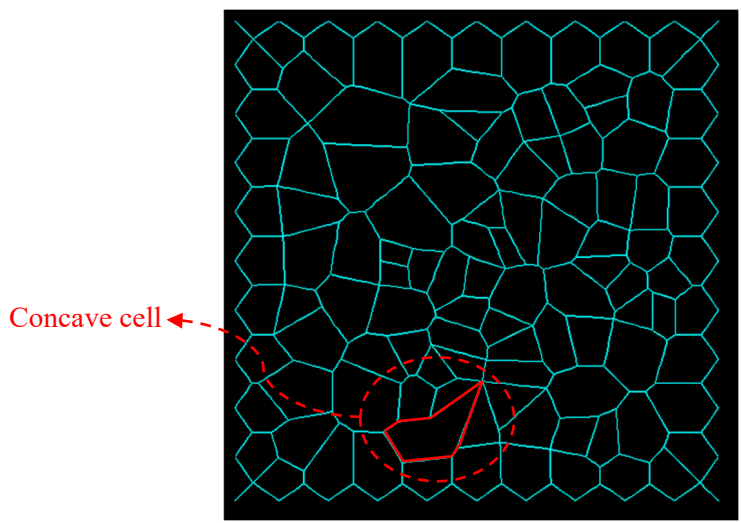

**Fig. 3.** An RVE example with a concave cell (image used for training CNNs).

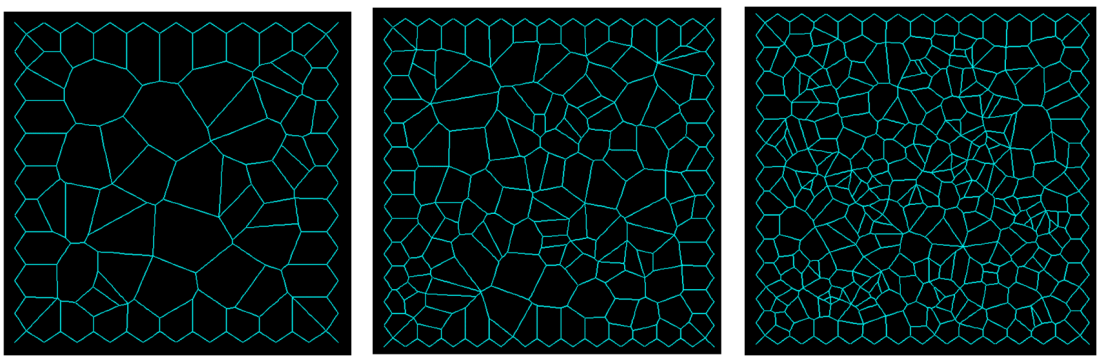

(a) Usual RVEs with random morphologies (most of the cases)

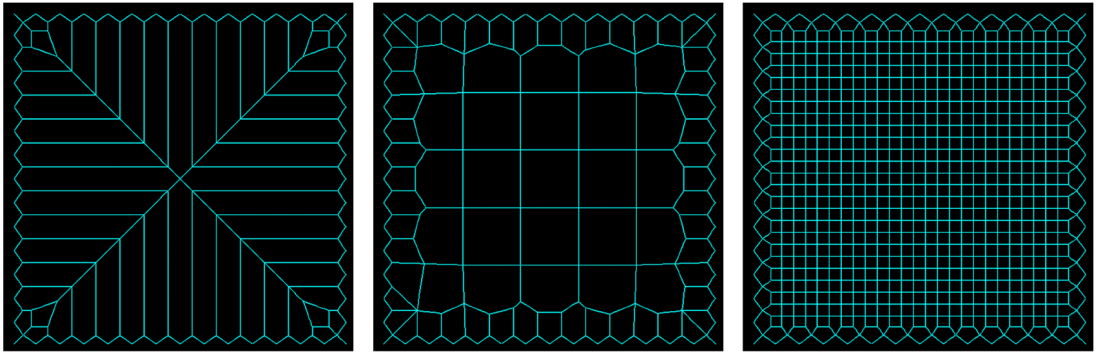
(b) Unusual RVEs with regularised morphologies (special cases)
**Fig. 4.** RVE examples (images used for training CNNs).

## 2.3. Convolutional neural networks

To date, multiple deep learning CNN models have been developed to facilitate and improve the feature extraction of images. One of these models is U-Net which was initially developed for the segmentation of biomedical images [36]. In the last couple of years, there have been many variations of U-Net introduced in different research areas, such as Unet++ [37], Attention U-Net [38], Residual Unet [39] and Nas U-Net [40], with each one modifying and improving the original architecture for their desired tasks. The general U-Net architecture is constructed of two paths: an encoder with the primary objective of finding features of different scales and semantic values from the input image and a decoder path responsible for upscaling produced feature maps to generate the final segmentation layers [41]. During the upscaling (up-convolutions), there is a concatenation from the last layer of the corresponding block in the encoder path. This concatenation helps the model regain some information that might be lost during downs sampling.

Another significant achievement in CNNs was the introduction of ResNets [42]. Originally it was expected that, as the number of layers increases, the model performance increases, but in practice, after a certain number of layers, the model performance would plummet due to the vanishing gradient issue [43]. Resnet solved this issue by introducing residual blocks and skip connections. Skip connections are the main differentiator between normal and residual blocks. These connections modify the block by adding the input of the block to the final layer output. One method to add input to the final layer output while maintaining the same dimension as output is through identity mapping, which is a 1 × 1 convolution. These skip connections help prevent the gradient from becoming too small, avoiding the model performance reduction. A simple residual block is shown in Fig. 5.

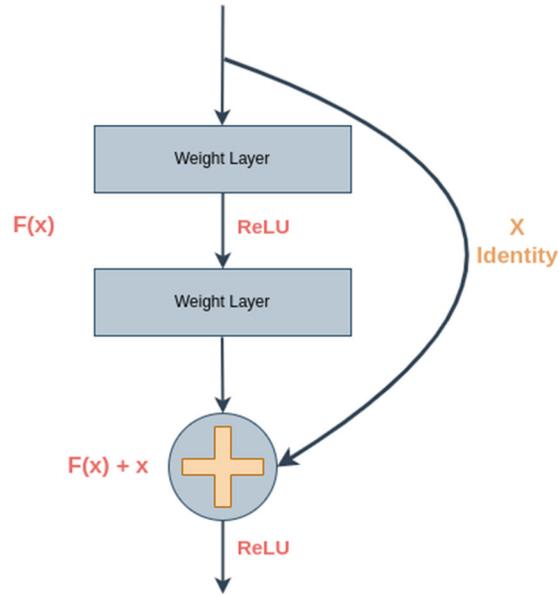

**Fig. 5.** Schematic representation of a residual block.

The model used in this work is a combination of the above-mentioned models originally proposed by [39] for road extraction from ariel images. Our model incorporates the downsampling part from the ResUnet model followed by a series of fully connected layers to generate a regressor model for the prediction of the Young's modulus. The skip connections and the residual block, as well as the model architecture, are shown in Fig. 6.

**Fig. 6.** Architecture of the designed CNN to establish the relationship between the porous meso-structure to the effective Young's modulus.

Given the nature of the predicted values, multiple loss functions were tested for optimal results. The Mean Absolute Error (MAE) provided significantly better results compared to other tested loss functions, i.e., Mean Squared Error (MSE) or Root Mean Squared Error (RMSE). The L1 loss function is given as

$$MAE = \frac{1}{n}\sum_{1}^{n}|Y_{actual} - Y_{predicted}| \qquad (4)$$

## 2.4. Results from CNN

We evaluate the proposed model performance for predicting the value of the effective Young's modulus, and the accepted prediction is less than a 10 % difference (see the outcome comparison for the validation case in section 2.1). The model was built on Pytorch 1.9, and the training was performed on a device with NVIDIA RTX 2080. In addition, augmentation (see section 2.2) was used for training. Finally, we performed further hyperparameter tuning to achieve the ultimate result. Since there is a wide distribution (i.e., a high range or scale) in the Young's modulus, the value of MAE does not provide a good general representation of the model regression performance. The relative error percentage is used to investigate the results to achieve a better metric.

The model achieved an average error of 5.92 %, which is well within the 10 % acceptable tolerance initially defined. The final error distribution is shown in Fig. 7. As can be seen, the model does a relatively good job predicting many images with a relatively low error rate while struggling with a few images. In fact, for the majority of the tested cases, the error for the prediction is less than 5 %. For some limited cases, we observe a rather high error. Observing the meso-structures related to those cases, we realized that the randomly generated patterns for these cases are accidentally way different compared to the majority of the dataset. The latter point explains the deviation in these cases, which can be easily overcome by providing more data for the network. Although we have used data augmentation, adding other physical constraints such as material symmetry and so on can be extremely helpful in increasing the predictions' accuracy for unseen situations [44, 45]. The last point remains to be shown in future developments.

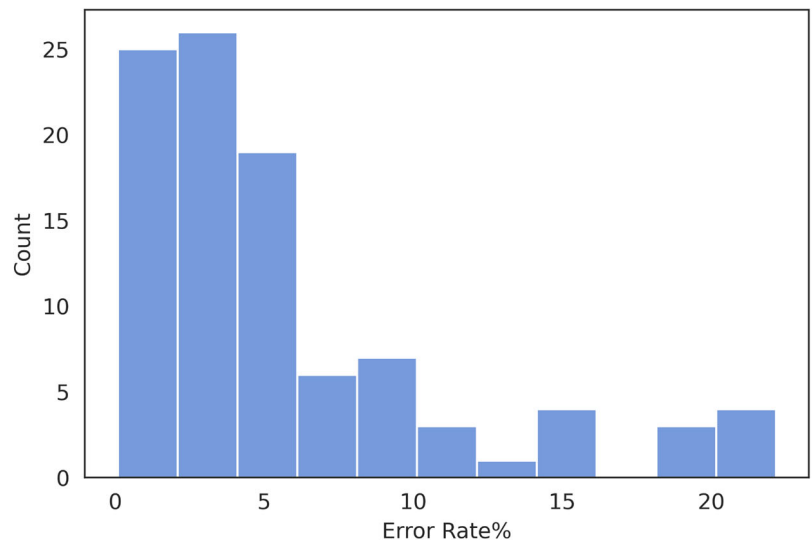

**Fig. 7.** Distribution of error rate on the test dataset.

## 3. Structural uncertainty analysis

This section employs the uncertain material parameters (3.1) caused by CNN prediction in the static bending analysis (3.2) of an FG beam containing three graded porous layers. The corresponding validation and discussion (3.3) are given to disclose the impact of material uncertainty on the structural performance.

### 3.1. Material uncertainty

As stated in section 2.4, the prediction from CNN model for closed-cell Aluminium foams has an averaged error range of ±5.92 %, resulting in the uncertain material properties. To incorporate this uncertainty into the structural assessment and evaluate its influence on the overall mechanical behaviour of FG porous beam, the fuzzy uncertainty [46-48] is adopted and associated to Young's modulus by defining the set of triangular fuzzy numbers $\tilde{E}(\alpha,\beta)$ with the double parametric form expressed as

*Young's modulus*

$$\tilde{E}(\alpha,\beta) = (e_1 - e_3)\alpha\beta + (e_3 - e_1)\beta + (e_2 - e_1)\alpha + e_1, \quad \alpha,\beta \in [0,1] \quad (5)$$

of which $e_2$ represents the FE RVE results regarded as the accurate data here, $e_1 = (1 - 5.92\%)e_2$ and $e_3 = (1 + 5.92\%)e_2$ correspond to the lower and upper limits of deep learning predictions, respectively. Meanwhile, other material parameters of metal foams remain deterministically linked to the relative density $\mu$:

*Shear modulus* [17] $\quad G(\mu) = G_0[(0.75 \times 0.4886)(1 + v_0)(0.5\mu^2 + 0.3\mu)] \quad (6)$

*Poisson's ratio* [49] $\quad v(\mu) = v_0 + 3(1 - 5v_0)(1 - v_0^2)(1 - \mu) \Big/ 2(7 - 5v_0) \quad (7)$

*Mass density* $\quad \rho(\mu) = \mu\rho_0 \quad (8)$

where $G_0 = 23.731$ GPa, $v_0 = 0.3$, and $\rho_0 = 2700$ kg/m³ for the cell wall matrix materials.

## 3.2. Bending analysis of FG porous beam

We then focus on the static bending behaviour of a typical FG porous beam containing a thick low-density layer and bounded by top and bottom high-density porous layers, as depicted in Fig. 8, where all the layers are assumed to be perfectly bonded together and made of closed-cell Aluminium foams. The relation between layer and beam thicknesses is defined by $h_H$ ($h_H$: high-density layer) $= \frac{1}{4}h_L$ ($h_L$: low-density layer) $= \frac{1}{6}h_T$ ($h_T$: total beam thickness). A Cartesian coordinate system $(x, z)$ is built on the mid-thickness plane: $x \in [0, L]$ ($L$ is the beam length), $z \in [-0.5h_T, 0.5h_T,]$, and the beam width is denoted by $b$. The beam bending deflection is measured under the application of a point load located at the mid span.

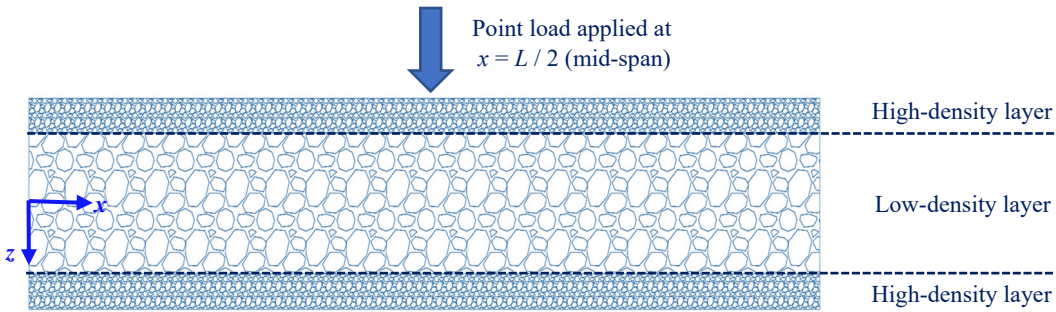

**Fig. 8.** Schematic drawing of an FG porous three-layer beam.

The relative densities of each layer are expressed as

$$\mu = \begin{cases} 13.81\ \%\ \text{(high-density layer)} \\ 6.92\ \%\ \ \text{(low-density layer)} \end{cases} \quad (9)$$

and the properties of both layers are uncertain with CNN predicted Young's moduli (based on Eq. (5)) written as

*Young's modulus of the high-density layer* ($\mu = 13.81\ \%$, $\alpha_1, \beta_1 \in [0,1]$)

$$\tilde{E}_H(\alpha_1, \beta_1) = (e_{1H} - e_{3H})\alpha_1\beta_1 + (e_{3H} - e_{1H})\beta_1 + (e_{2H} - e_{1H})\alpha_1 + e_{1H} \quad (10)$$

*Young's modulus of the low-density layer* ($\mu = 6.92\ \%$, $\alpha_2, \beta_2 \in [0,1]$)

$$\tilde{E}_L(\alpha_2, \beta_2) = (e_{1L} - e_{3L})\alpha_2\beta_2 + (e_{3L} - e_{1L})\beta_2 + (e_{2L} - e_{1L})\alpha_2 + e_{1L} \quad (11)$$

where $e_{2H} = 1693.92$ MPa, $e_{1H} = (1 - 5.92\%)e_{2H}$, and $e_{3H} = (1 + 5.92\%)e_{2H}$. Similarly, $e_{2L} = 865.98$ MPa, $e_{1L} = (1 - 5.92\%)e_{2L}$, and $e_{3L} = (1 + 5.92\%)e_{2L}$. Note that 1693.92 MPa and 865.98 MPa originate from the averaged compressive stiffnesses (Eq. (3)) of RVE sets featured by relative density ranges of (13.61 % ~ 14.02 %, mean relative density = 13.81 %) and (6.77 % ~ 7.08 %, mean relative density = 6.92 %), respectively, both of which include 100 cases with various random porosity distributions shown in Fig. 9 for examples. Moreover, other material properties of porous layers are deterministic and computed via Eqs. (6)-(8).

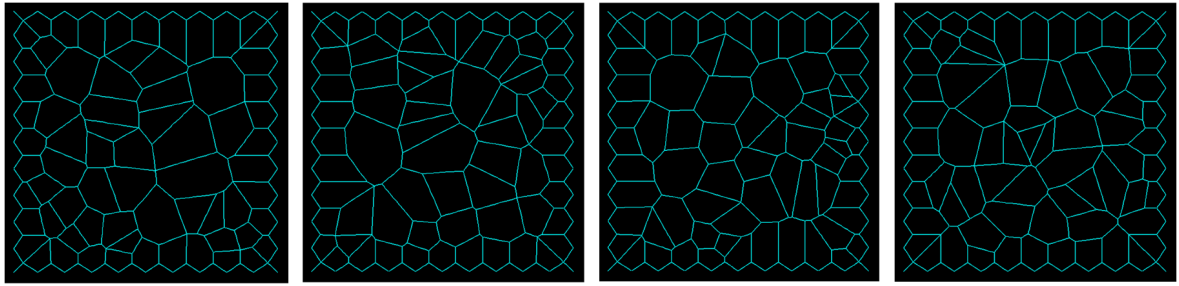

(a) Low-density layer ($\mu = \sim 6.92\ \%$)

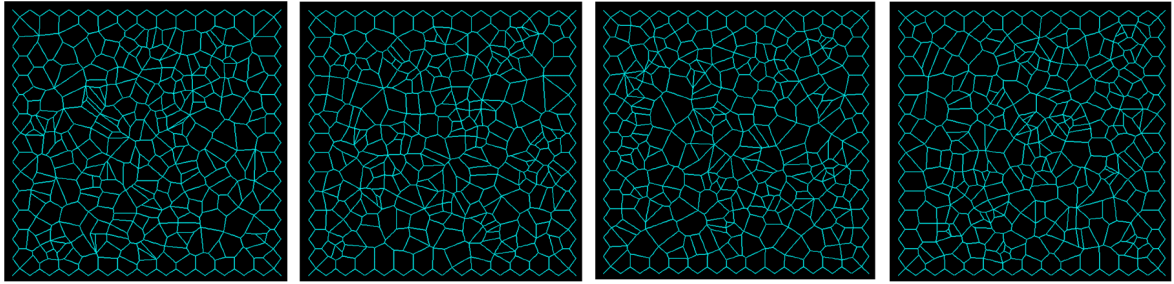

(b) High-density layer ($\mu = \sim 13.81\ \%$)

**Fig. 9.** RVE examples for porous layers (images used for training CNNs).

The bending analysis of the described FG porous beam is performed within the framework of Timoshenko beam theory:

$$\begin{cases} u^*(x,z) = u(x) + z\varphi(x) \\ w^*(x,z) = w(x) \end{cases} \quad (12)$$

in which $u^*(x,z)$ and $w^*(x,z)$ represent the axial and transverse beam displacements, $u(x)$ and $w(x)$ denote the corresponding displacement components on the mid-thickness plane, and $\varphi(x)$ reads the cross section rotation. Consequently, the normal and shear strains and stresses are expressed as [17, 50, 51]

Normal strain $\quad\quad\quad\quad\quad\quad\quad \varepsilon_{xx} = \partial u(x)/\partial x + z\left[\partial \varphi(x)/\partial x\right] \quad\quad (13)$

Shear strain $\quad\quad\quad\quad\quad\quad\quad \gamma_{xz} = \partial w(x)/\partial x + \varphi(x) \quad\quad (14)$

Normal stress $\quad\quad\quad\quad\quad\quad\quad \sigma_{xx} = \left(\dfrac{\tilde{E}}{1-v^2}\right)\varepsilon_{xx} \quad\quad (15)$

Shear stress $\quad\quad\quad\quad\quad\quad\quad \sigma_{xz} = G\gamma_{xz} \quad\quad (16)$

The beam strain energy and external load work can be subsequently obtained by [16]

Strain energy $\quad\quad\quad\quad \Lambda_S = \dfrac{b}{2}\int_0^L \int_{-0.5h_T}^{0.5h_T} (\varepsilon_{xx}\sigma_{xx} + \gamma_{xz}\sigma_{xz})dzdx \quad\quad (17)$

External load work $\quad\quad\quad\quad \Lambda_P = Pw(L/2) \quad\quad (18)$

where $P$ is an external point load applied on the mid-length point of the beam ($x = L/2$).

Therefore, the total energy functional reads $\Psi = \Lambda_S - \Lambda_P$, which can be solved with Ritz method [14] to determine the mid-span beam bending deflection $\Delta$. The beam boundary conditions are linked to the Ritz trial functions [52] (total number of polynomial terms = 10), and hinged-hinged (H-H) end supports are adopted in this research.

### 3.3. Validation and discussion on FG porous beam

A validation study is conducted first to identify the accuracy of the beam formulations in section 3.2, as given in Table 1. The following parameter settings are employed in both validation and the subsequent structural uncertainty analysis: $P = 100$ N, $h_T = 0.1$ m, $b = 0.1$ m, $L = 2$ m (slenderness ratio $L/h_T = 20$), H-H beam. The present results are compared to the theoretical and numerical outcomes, computed via Eq. (19) and commercial software ANSYS using 3000 solid elements (SOLID186). As can be seen, a high level of agreement demonstrates the validity of the present formulations.

For homogeneous beam $\quad\quad\quad\quad \Delta = \dfrac{PL^3}{48E'I'} \quad\quad (19)$

where $E'$ is based on Eq. (15) and $I' = bh_T^3/12$.

**Table 1**
Validation on mid-span deflections (mm).

| Eq. (19) | ANSYS | Present result |
|---|---|---|
| Homogeneous porous beam (made of high-density layer materials, deterministic) | | |
| 1.137 | 1.153 | 1.146 |
| Homogeneous porous beam (made of low-density layer materials, deterministic) | | |
| 2.231 | 2.266 | 2.251 |
| FG porous beam ($\alpha_1 = 1, \beta_1 = 1, \alpha_2 = 1, \beta_2 = 1$) | | |
| | 1.358 | 1.344 |
| FG porous beam ($\alpha_1 = 0, \beta_1 = 1, \alpha_2 = 0, \beta_2 = 1$) | | |
| | 1.283 | 1.270 |
| FG porous beam ($\alpha_1 = 0, \beta_1 = 0, \alpha_2 = 0, \beta_2 = 0$) | | |
| | 1.441 | 1.428 |

The structural uncertainty of the target FG porous beam is examined on the basis of varying mid-span deflections subject to a fixed point load (see Fig. 8), considering the indeterministic CNN predicted Young's moduli with triangular fuzzy numbers. Tables 2, 3, and 4 tabulate the deflection results in terms of the uncertainties induced from high-density layer, low-density layer, and both layers, while the corresponding upper and lower bounds are depicted in Figs. 10 (a), 10 (b), and 10 (c), respectively. It is found that the fuzziness of high-density layer produces a larger deflection range (1.283 ~ 1.412, $\alpha_1 = 0$) than that related with low-density layer (1.331 ~ 1.358, $\alpha_2 = 0$), as evidenced by the higher upper bound and lower lower bound. In the meanwhile, when accounting for the uncertainties of both layers simultaneously, the deflection range is further extended with distinctly separated limits (1.270 ~ 1.428, $\alpha_1 = \alpha_2 = 0$). This indicates that the indeterministic properties from top and bottom high-density layers impact the porous beam stiffness more significantly than the low-density layer one, despite the obvious difference in the layer thickness, i.e., the high-density layer is thinner than its low-density counterpart. The design of this FG porous beam is rather similar to a sandwich configuration, and hence the strong facings contribute to the overall structural rigidity in a more effective manner than the soft lightweight core.

**Table 2**
Mid-span deflections (mm) considering high-density layer uncertainty ($\alpha_2 = 1$, $\beta_2 = 1$).

| $\alpha_1$ | $\beta_1$ | | | | |
|---|---|---|---|---|---|
| | 0 | 0.25 | 0.5 | 0.75 | 1 |
| 0   | 1.412 | 1.378 |       | 1.313 | 1.283 |
| 0.1 | 1.405 | 1.374 |       | 1.316 | 1.288 |
| 0.2 | 1.398 | 1.371 |       | 1.319 | 1.294 |
| 0.3 | 1.391 | 1.367 |       | 1.322 | 1.300 |
| 0.4 | 1.384 | 1.364 |       | 1.325 | 1.307 |
| 0.5 | 1.378 | 1.361 | 1.344 | 1.328 | 1.313 |
| 0.6 | 1.371 | 1.357 |       | 1.332 | 1.319 |
| 0.7 | 1.364 | 1.354 |       | 1.335 | 1.325 |
| 0.8 | 1.357 | 1.351 |       | 1.338 | 1.332 |
| 0.9 | 1.351 | 1.348 |       | 1.341 | 1.338 |
| 1   | 1.344 | 1.344 |       | 1.344 | 1.344 |

**Table 3**
Mid-span deflections (mm) considering low-density layer uncertainty ($\alpha_1 = 1$, $\beta_1 = 1$).

| $\alpha_2$ | $\beta_2$ | | | | |
|---|---|---|---|---|---|
| | 0 | 0.25 | 0.5 | 0.75 | 1 |
| 0   | 1.358 | 1.351 |       | 1.337 | 1.331 |
| 0.1 | 1.357 | 1.351 |       | 1.338 | 1.332 |
| 0.2 | 1.356 | 1.350 |       | 1.339 | 1.333 |
| 0.3 | 1.354 | 1.349 |       | 1.339 | 1.335 |
| 0.4 | 1.353 | 1.349 |       | 1.340 | 1.336 |
| 0.5 | 1.351 | 1.348 | 1.344 | 1.341 | 1.337 |
| 0.6 | 1.350 | 1.347 |       | 1.342 | 1.339 |
| 0.7 | 1.349 | 1.346 |       | 1.342 | 1.340 |
| 0.8 | 1.347 | 1.346 |       | 1.343 | 1.342 |
| 0.9 | 1.346 | 1.345 |       | 1.344 | 1.343 |
| 1   | 1.344 | 1.344 |       | 1.344 | 1.344 |

**Table 4**
Mid-span deflections (mm) considering uncertainties of both layers.

| $\alpha_1, \alpha_2$ | $\beta_1, \beta_2$ | | | | |
|---|---|---|---|---|---|
| | 0 | 0.25 | 0.5 | 0.75 | 1 |
| **0**   | 1.428 | 1.385 |       | 1.306 | 1.270 |
| **0.1** | 1.419 | 1.381 |       | 1.310 | 1.277 |
| **0.2** | 1.410 | 1.377 |       | 1.314 | 1.284 |
| **0.3** | 1.402 | 1.372 |       | 1.317 | 1.291 |
| **0.4** | 1.393 | 1.368 |       | 1.321 | 1.299 |
| **0.5** | 1.385 | 1.364 | 1.344 | 1.325 | 1.306 |
| **0.6** | 1.377 | 1.360 |       | 1.329 | 1.314 |
| **0.7** | 1.368 | 1.356 |       | 1.333 | 1.321 |
| **0.8** | 1.360 | 1.352 |       | 1.337 | 1.329 |
| **0.9** | 1.352 | 1.348 |       | 1.340 | 1.337 |
| **1**   | 1.344 | 1.344 |       | 1.344 | 1.344 |

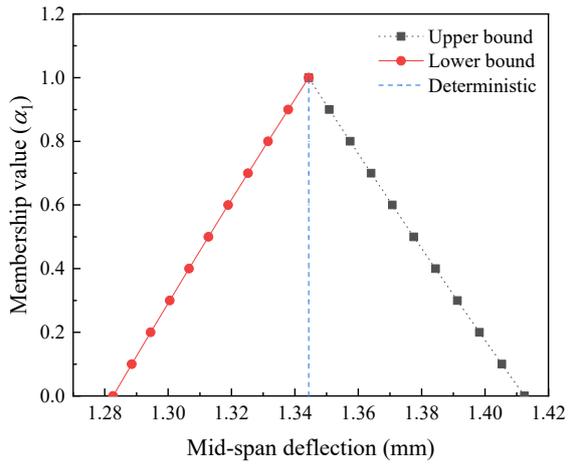

(a) High-density layer uncertainty

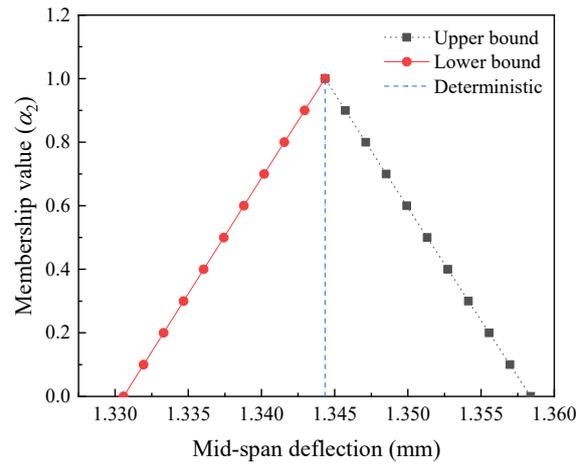

(b) Low-density layer uncertainty

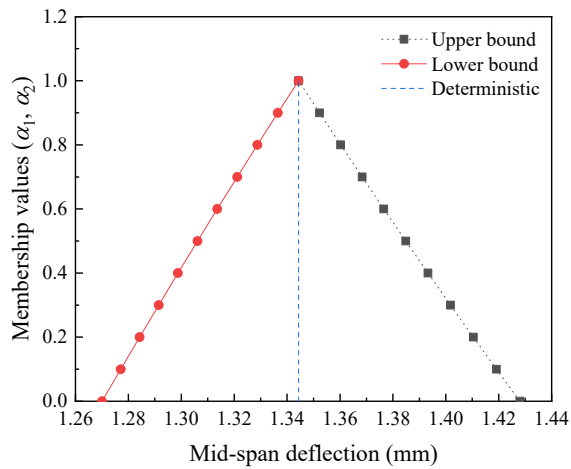

(c) Uncertainties of both layers

**Fig. 10.** Upper and lower bounds of uncertain mid-span deflections.

## 4. Conclusions

This chapter proposes a novel framework for AI enhanced FE multiscale modelling of FG porous structures. A database contains FE RVEs featured with various cellular morphologies and corresponding to the practical closed-cell Aluminium foams. The deep learning CNN model is designed and trained to predict the effective Young's modulus of RVEs simply based on their images. The structural uncertainty originated from CNN result errors (±5.92 %) is revealed with a bending assessment on the FG porous three-layer beam. Our findings show that the developed method has a great potential to quickly and conveniently evaluate the porous structural performance according to the mesoscopic characteristics of foam geometries extracted from images, bypassing the detailed mechanical modelling. The consequent behaviour uncertainty can also be quantitatively measured.

In the study here, 2D images provide a relatively easy path for the establishment of the relationship between foam morphologies and properties. However, 3D analysis normally gives more in-depth insights into micro- and macro-responses of cellular composites and the connection may also be captured by CNN models. We will include this into our future research.

**Acknowledgments**

The author Chen acknowledges the financial support from the Australian Research Council (ARC DECRA DE220100876).

**References**

[1] Alantali A, Alia RA, Umer R, Cantwell WJ. Energy absorption in aluminium honeycomb cores reinforced with carbon fibre reinforced plastic tubes. Journal of Sandwich Structures & Materials. 2017;21:2801-15.
[2] Wang C, Cheng H, Hong C, Zhang X, Zeng T. Lightweight chopped carbon fibre reinforced silica-phenolic resin aerogel nanocomposite: Facile preparation, properties and application to thermal protection. Composites Part A: Applied Science and Manufacturing. 2018;112:81-90.
[3] Ahmad H, Markina AA, Porotnikov MV, Ahmad F. A review of carbon fiber materials in automotive industry. IOP Conference Series: Materials Science and Engineering. 2020;971:032011.
[4] Jha K, Yeswanth IVS, Manish D, Tyagi YK. Structural and Modal Analysis of PEEK-CF Composite for Aircraft Wing. In: Parey A, Kumar R, Singh M, editors. Recent Trends in Engineering Design. Singapore: Springer Singapore; 2021. p. 101-12.
[5] Norkhairunnisa M, Chai Hua T, Sapuan SM, Ilyas RA. Evolution of Aerospace Composite Materials. In: Mazlan N, Sapuan SM, Ilyas RA, editors. Advanced Composites in Aerospace Engineering Applications. Cham: Springer International Publishing; 2022. p. 367-85.
[6] Popescu M, Reiter L, Liew A, Van Mele T, Flatt RJ, Block P. Building in Concrete with an Ultra-lightweight Knitted Stay-in-place Formwork: Prototype of a Concrete Shell Bridge. Structures. 2018;14:322-32.
[7] Bechert S, Aldinger L, Wood D, Knippers J, Menges A. Urbach Tower: Integrative structural design of a lightweight structure made of self-shaped curved cross-laminated timber. Structures. 2021;33:3667-81.

[8] Liu H, Zhang ET, Wang G, Ng BF. In-plane crushing behavior and energy absorption of a novel graded honeycomb from hierarchical architecture. International Journal of Mechanical Sciences. 2022;221:107202.
[9] Korkmaz ME, Gupta MK, Robak G, Moj K, Krolczyk GM, Kuntoğlu M. Development of lattice structure with selective laser melting process: A state of the art on properties, future trends and challenges. Journal of Manufacturing Processes. 2022;81:1040-63.
[10] García-Moreno F. Commercial applications of metal foams: their properties and production. Materials. 2016;9:85.
[11] Chen D, Yang J, Schneider J, Kitipornchai S, Zhang L. Impact response of inclined self-weighted functionally graded porous beams reinforced by graphene platelets. Thin-Walled Structures. 2022;179:109501.
[12] Gao K, Gao W, Chen D, Yang J. Nonlinear free vibration of functionally graded graphene platelets reinforced porous nanocomposite plates resting on elastic foundation. Composite Structures. 2018;204:831-46.
[13] Wu H, Yang J, Kitipornchai S. Mechanical analysis of functionally graded porous structures: A review. International Journal of Structural Stability and Dynamics. 2020;20:2041015.
[14] Chen D, Yang J, Kitipornchai S. Buckling and bending analyses of a novel functionally graded porous plate using Chebyshev-Ritz method. Archives of Civil and Mechanical Engineering. 2019;19:157-70.
[15] Chen D, Kitipornchai S, Yang J. Dynamic response and energy absorption of functionally graded porous structures. Materials & Design. 2018;140:473-87.
[16] Chen D, Yang J, Kitipornchai S. Elastic buckling and static bending of shear deformable functionally graded porous beam. Composite Structures. 2015;133:54-61.
[17] Chen D, Rezaei S, Rosendahl PL, Xu BX, Schneider J. Multiscale modelling of functionally graded porous beams: Buckling and vibration analyses. Engineering Structures. 2022;266:114568.
[18] Huang JS, Liew JX, Ademiloye AS, Liew KM. Artificial Intelligence in Materials Modeling and Design. Archives of Computational Methods in Engineering. 2021;28:3399-413.
[19] Zhang C, Lu Y. Study on artificial intelligence: The state of the art and future prospects. Journal of Industrial Information Integration. 2021;23:100224.
[20] Pan Y, Zhang L. Roles of artificial intelligence in construction engineering and management: A critical review and future trends. Automation in Construction. 2021;122:103517.
[21] Wang C, Song LH, Fan JS. End-to-End Structural analysis in civil engineering based on deep learning. Automation in Construction. 2022;138:104255.
[22] Baduge SK, Thilakarathna S, Perera JS, Arashpour M, Sharafi P, Teodosio B et al. Artificial intelligence and smart vision for building and construction 4.0: Machine and deep learning methods and applications. Automation in Construction. 2022;141:104440.
[23] Jacobsen EL, Teizer J. Deep Learning in Construction: Review of Applications and Potential Avenues. Journal of Computing in Civil Engineering. 2022;36:03121001.
[24] Perez-Ramirez CA, Amezquita-Sanchez JP, Valtierra-Rodriguez M, Adeli H, Dominguez-Gonzalez A, Romero-Troncoso RJ. Recurrent neural network model with Bayesian training and mutual information for response prediction of large buildings. Engineering Structures. 2019;178:603-15.
[25] Sony S, Dunphy K, Sadhu A, Capretz M. A systematic review of convolutional neural network-based structural condition assessment techniques. Engineering Structures. 2021;226:111347.
[26] Batra R, Song L, Ramprasad R. Emerging materials intelligence ecosystems propelled by machine learning. Nature Reviews Materials. 2021;6:655-78.
[27] Mianroodi JR, Rezaei S, Siboni NH, Xu BX, Raabe D. Lossless multi-scale constitutive elastic relations with artificial intelligence. npj Computational Materials. 2022;8:67.
[28] Zhou X, Yang Y, Bharech S, Lin B, Schröder J, Xu BX. 3D-multilayer simulation of microstructure and mechanical properties of porous materials by selective sintering. GAMM-Mitteilungen. 2021;44:e202100017.
[29] Javili A, Chatzigeorgiou G, Steinmann P. Computational homogenization in magneto-mechanics. International Journal of Solids and Structures. 2013;50:4197-216.

[30] Yang Y, Fathidoost M, Oyedeji TD, Bondi P, Zhou X, Egger H et al. A diffuse-interface model of anisotropic interface thermal conductivity and its application in thermal homogenization of composites. Scripta Materialia. 2022;212:114537.

[31] Liang M, Li Z, Lu F, Li X. Theoretical and numerical investigation of blast responses of continuous-density graded cellular materials. Composite Structures. 2017;164:170-9.

[32] Li D, Dong L, Yin J, Lakes RS. Negative Poisson's ratio in 2D Voronoi cellular solids by biaxial compression: a numerical study. Journal of Materials Science. 2016;51:7029-37.

[33] Verma KS, Panthi SK, Mondal D. Compressive deformation behavior of closed cell LM-13 aluminum alloy foam using finite element analysis. Materials Today: Proceedings. 2020;28:1073-7.

[34] Kurniati EO, Dirgantara T, Gunawan L, Jusuf A. Meso-modeling of Closed-Cell Aluminum Foam Under Compression Loading. Advances in Lightweight Materials and Structures: Springer; 2020. p. 3-17.

[35] Luo G, Xue P, Li Y. Experimental investigation on the yield behavior of metal foam under shear-compression combined loading. Science China Technological Sciences. 2021;64:1412-22.

[36] Ronneberger O, Fischer P, Brox T. U-Net: Convolutional Networks for Biomedical Image Segmentation. ArXiv. 2015;arXiv:1505.04597

[37] Zhou Z, Rahman Siddiquee MM, Tajbakhsh N, Liang J. UNet++: A Nested U-Net Architecture for Medical Image Segmentation. In: Stoyanov D, Taylor Z, Carneiro G, Syeda-Mahmood T, Martel A, Maier-Hein L, et al., editors. Deep Learning in Medical Image Analysis and Multimodal Learning for Clinical Decision Support. Cham: Springer International Publishing; 2018. p. 3-11.

[38] Oktay O, Schlemper J, Folgoc LL, Lee M, Heinrich M, Misawa K et al. Attention u-net: Learning where to look for the pancreas. arXiv preprint arXiv:180403999. 2018.

[39] Zhang Z, Liu Q, Wang Y. Road Extraction by Deep Residual U-Net. IEEE Geoscience and Remote Sensing Letters. 2018;15:749-53.

[40] Weng Y, Zhou T, Li Y, Qiu X. NAS-Unet: Neural Architecture Search for Medical Image Segmentation. IEEE Access. 2019;7:44247-57.

[41] Hou Y, Liu Z, Zhang T, Li Y. C-UNet: Complement UNet for Remote Sensing Road Extraction. Sensors. 2021;21:2153.

[42] He K, Zhang X, Ren S, Sun J. Deep residual learning for image recognition. ArXiv. 2015;arXiv:1512.03385

[43] Basodi S, Ji C, Zhang H, Pan Y. Gradient amplification: An efficient way to train deep neural networks. Big Data Mining and Analytics. 2020;3:196-207.

[44] Xu K, Huang DZ, Darve E. Learning constitutive relations using symmetric positive definite neural networks. Journal of Computational Physics. 2021;428:110072.

[45] Rezaei S, Harandi A, Moeineddin A, Xu BX, Reese S. A mixed formulation for physics-informed neural networks as a potential solver for engineering problems in heterogeneous domains: Comparison with finite element method. Computer Methods in Applied Mechanics and Engineering. 2022;401:115616.

[46] Jena SK, Chakraverty S, Jena RM. Propagation of uncertainty in free vibration of Euler–Bernoulli nanobeam. Journal of the Brazilian Society of Mechanical Sciences and Engineering. 2019;41:1-18.

[47] Jena SK. Vibrations of functionally graded structure with material uncertainties. Modeling and Computation in Vibration Problems. 2021;2:8-1.

[48] Jena SK, Chakraverty S, Malikan M. Implementation of non-probabilistic methods for stability analysis of nonlocal beam with structural uncertainties. Engineering with Computers. 2021;37:2957-69.

[49] Gregorová E, Pabst W, Uhlířová T, Nečina V, Veselý M, Sedlářová I. Processing, microstructure and elastic properties of mullite-based ceramic foams prepared by direct foaming with wheat flour. Journal of the European Ceramic Society. 2016;36:109-20.

[50] Rosendahl PL, Weißgraeber P. Modeling snow slab avalanches caused by weak-layer failure–Part 1: Slabs on compliant and collapsible weak layers. The Cryosphere. 2020;14:115-30.

[51] Rosendahl PL, Weißgraeber P. Modeling snow slab avalanches caused by weak-layer failure–Part 2: Coupled mixed-mode criterion for skier-triggered anticracks. The Cryosphere. 2020;14:131-45.

[52] Kitipornchai S, Chen D, Yang J. Free vibration and elastic buckling of functionally graded porous beams reinforced by graphene platelets. Materials & Design. 2017;116:656-65.